\pdfoutput=1

\documentclass[11pt]{article}

\usepackage{ACL2023}

\usepackage{times}
\usepackage{latexsym}
\usepackage{comment}
\usepackage{booktabs}
\usepackage[T1]{fontenc}

\usepackage[utf8]{inputenc}

\usepackage{microtype}

\usepackage{inconsolata}

\usepackage{graphicx}
\usepackage[capitalise]{cleveref}
\usepackage{multirow}
\usepackage{longtable}

\usepackage[edges]{forest}
\tikzset{%
    parent/.style =          {align=center,text width=2cm,rounded corners=3pt},
    child/.style =           {align=center,text width=1cm,rounded corners=3pt},
    grandchild/.style =      {align=center,text width=2cm,rounded corners=3pt},
    greatgrandchild/.style = {align=left,text width=8.5cm,rounded corners=3pt}
}

\usepackage{tikz}
\usepackage{tikz-qtree}
\usetikzlibrary{trees} 

\usepackage{pgf,tikz}
\usepackage{forest}
\usetikzlibrary{trees,positioning,shapes,shadows,arrows.meta}

\newcommand{\xmark}{%
\tikz[scale=0.23, color=red] {
    \draw[line width=0.7,line cap=round] (0,0) to [bend left=6] (1,1);
    \draw[line width=0.7,line cap=round] (0.2,0.95) to [bend right=3] (0.8,0.05);
}}
\newcommand{\cmark}{%
\tikz[scale=0.23, color=green] {
    \draw[line width=0.7,line cap=round] (0.25,0) to [bend left=10] (1,1);
    \draw[line width=0.8,line cap=round] (0,0.35) to [bend right=1] (0.23,0);
}}

\newcommand*{\affaddr}[1]{#1}
\newcommand*{\affmark}[1][*]{\textsuperscript{#1}}
\newcommand*{\email}[1]{\texttt{#1}}
\author{
Vipula Rawte\affmark[1]\thanks{corresponding author}, Amit Sheth\affmark[1], Amitava Das\affmark[1]  \\
\affaddr{\affmark[1]AI Institute, University of South Carolina, USA}\\
\email{\{vrawte\}@mailbox.sc.edu}
}

\title{A Survey of Hallucination in ``Large'' Foundation Models}

\begin{document}
\maketitle
\begin{abstract}

Hallucination in a foundation model (FM) refers to the generation of content that strays from factual reality or includes fabricated information. This survey paper provides an extensive overview of recent efforts that aim to identify, elucidate, and tackle the problem of hallucination, with a particular focus on ``Large'' Foundation Models (LFMs). The paper classifies various types of hallucination phenomena that are specific to LFMs and establishes evaluation criteria for assessing the extent of hallucination. It also examines existing strategies for mitigating hallucination in LFMs and discusses potential directions for future research in this area. Essentially, the paper offers a comprehensive examination of the challenges and solutions related to hallucination in LFMs.

\end{abstract}

\section{Introduction}

Foundation Models (FMs), exemplified by GPT-3 \cite{brown2020language} and Stable Diffusion \cite{rombach2022high}, marks the commencement of a novel era in the realm of machine learning and generative artificial intelligence. Researchers introduced the term \textbf{``foundation model''} to describe machine learning models that are trained on extensive, diverse, and unlabeled data, enabling them to proficiently handle a wide array of general tasks. These tasks encompass language comprehension, text and image generation, and natural language conversation.

\subsection{What is a Foundation Model}

Foundation models refer to massive AI models trained on extensive volumes of unlabeled data, typically through self-supervised learning. This training approach yields versatile models capable of excelling in a diverse range of tasks, including image classification, natural language processing, and question-answering, achieving remarkable levels of accuracy.

These models excel in tasks involving generative abilities and human interaction, such as generating marketing content or producing intricate artwork based on minimal prompts. However, adapting and implementing these models for enterprise applications can present certain difficulties \cite{bommasani2021opportunities}.

\subsection{What is Hallucination in Foundation Model?}

Hallucination in the context of a foundation model refers to a situation where the model generates content that is not based on factual or accurate information. Hallucination can occur when the model produces text that includes details, facts, or claims that are fictional, misleading, or entirely fabricated, rather than providing reliable and truthful information.

This issue arises due to the model's ability to generate plausible-sounding text based on patterns it has learned from its training data, even if the generated content does not align with reality. Hallucination can be unintentional and may result from various factors, including biases in the training data, the model's lack of access to real-time or up-to-date information, or the inherent limitations of the model in comprehending and generating contextually accurate responses.

\begin{figure}[!ht]
\includegraphics[width=8cm]{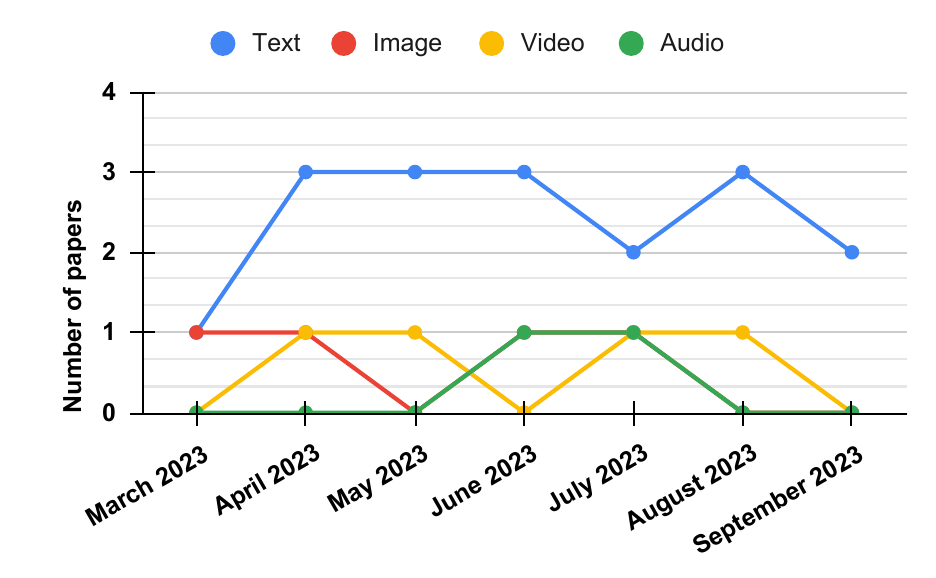}
\caption{The evolution of ``hallucination'' papers for Large Foundation Models (LFMs) from March 2023 to September 2023.}
\label{fig:paper-stats}
\end{figure}

Addressing hallucination in foundation models and LLMs is crucial, especially in applications where factual accuracy is paramount, such as journalism, healthcare, and legal contexts. Researchers and developers are actively working on techniques to mitigate hallucinations and improve the reliability and trustworthiness of these models. With the recent rise in this problem \cref{fig:paper-stats}, it has become even more critical to address them.

\subsection{Why this survey?}

In recent times, there has been a significant surge of interest in LFMs within both academic and industrial sectors. Additionally, one of their main challenges is \textit{hallucination}. The survey in \cite{ji2023survey} describes hallucination in natural language generation. In the era of \textbf{large} models, \cite{zhang2023siren} have done another great timely survey studying hallucination in LLMs. However, besides not only in LLMs, the problem of hallucination also exists in other foundation models such as image, video, and audio as well. Thus, in this paper, we do the first comprehensive survey of hallucination across all major modalities of foundation models.

\subsubsection{Our contributions}
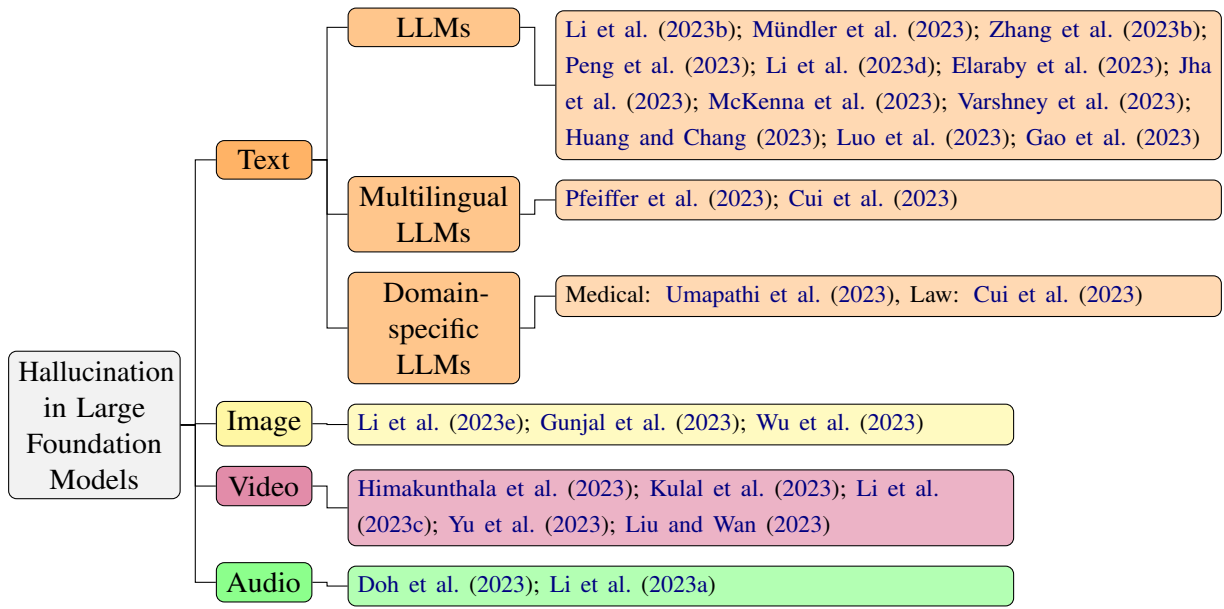
\begin{figure*}[!ht]
    \begin{center}
            \begin{forest}
                for tree={
                    forked edges,
                    grow'=0,
                    draw,
                    rounded corners,
                    node options={align=center,},
                    text width=2.7cm,
                },
                [Hallucination
in Large Foundation Models, fill=gray!10, parent
                    [Text, for tree={fill=orange!60, child}
                        [LLMs, fill=orange!45, grandchild[\footnotesize{\citet{li2023helma,mündler2023selfcontradictory,zhang2023mitigating,peng2023check,li2023chain,elaraby2023halo,jha2023dehallucinating,mckenna2023sources,varshney2023stitch,huang2023citation,luo2023zero,gao2023rarr}}, fill=orange!30, greatgrandchild
                            ]
                        ]
                        [Multilingual LLMs, fill=orange!45, grandchild
                        [\footnotesize{\citet{pfeiffer2023mmt5,cui2023chatlaw}}, fill=orange!30, greatgrandchild
                            ]
                        ]
                        [Domain-specific LLMs, fill=orange!45, grandchild
                        [\footnotesize{Medical: \citet{umapathi2023med}, Law: \citet{cui2023chatlaw}}, fill=orange!30, greatgrandchild
                            ]
                        ]
                    ]
                    [Image, for tree={fill=yellow!45,child}
                        [\footnotesize {\citet{li2023evaluating,gunjal2023detecting,wu2023hallucination}}, fill=yellow!30, greatgrandchild]
                    ]
                    [Video, for tree={fill=purple!45, child}
                        [\footnotesize {\citet{himakunthala2023let,kulal2023putting,li2023videochat,yu2023deficiency,liu2023models}}, fill=purple!30, greatgrandchild]
                    ]
                    [Audio, for tree={fill=green!45, child}
                        [\footnotesize {\citet{doh2023lp,li2023audio}}, fill=green!30, greatgrandchild]
                    ]
                ]
            \end{forest}
    \end{center}

\caption{Taxonomy for Hallucination in Large Foundation Models}
\label{fig:taxonomy}
\end{figure*}

The contributions of this survey paper are as follows:

\begin{enumerate}
  \item We succinctly categorize the existing works in the area of hallucination in LFMs, as shown in \cref{fig:taxonomy}.
  \item We offer an extensive examination of large foundation models (LFMs) in \cref{sec:text,sec:image,sec:video,sec:audio}. 
  \item We cover all the important aspects such as i. detection, ii. mitigation, iii. tasks, iv. datasets, and v. evaluation metrics, given in \cref{tab:big-table}.
  \item We finally also provide our views and possible future direction in this area. We will regularly update the associated open-source resources, available for access at \url{https://github.com/vr25/hallucination-foundation-model-survey}
\end{enumerate}

\subsubsection{Classification of Hallucination}
As shown in \cref{fig:taxonomy}, we broadly classify the LFMs into \textbf{four} types as follows: i. Text, ii. Image, iii. video, and iv. Audio. 

The paper follows the following structure.
Based on the above classification, we describe the hallucination and mitigation techniques for all four modalities in: i. text (\cref{sec:text}), ii. image (\cref{sec:image}), iii. video (\cref{sec:video}), and iv. audio (\cref{sec:audio}). In \cref{sec:good-hal}, we briefly discuss how hallucinations are NOT always bad, and hence, in the creative domain, they can be well-suited to producing artwork. Finally, we give some possible future directions for addressing this issue along with a conclusion in \cref{sec:conclusion}.

\section{Hallucination in Large Language Models} \label{sec:text}

\begin{figure*}[!ht]
\includegraphics[width=15cm]{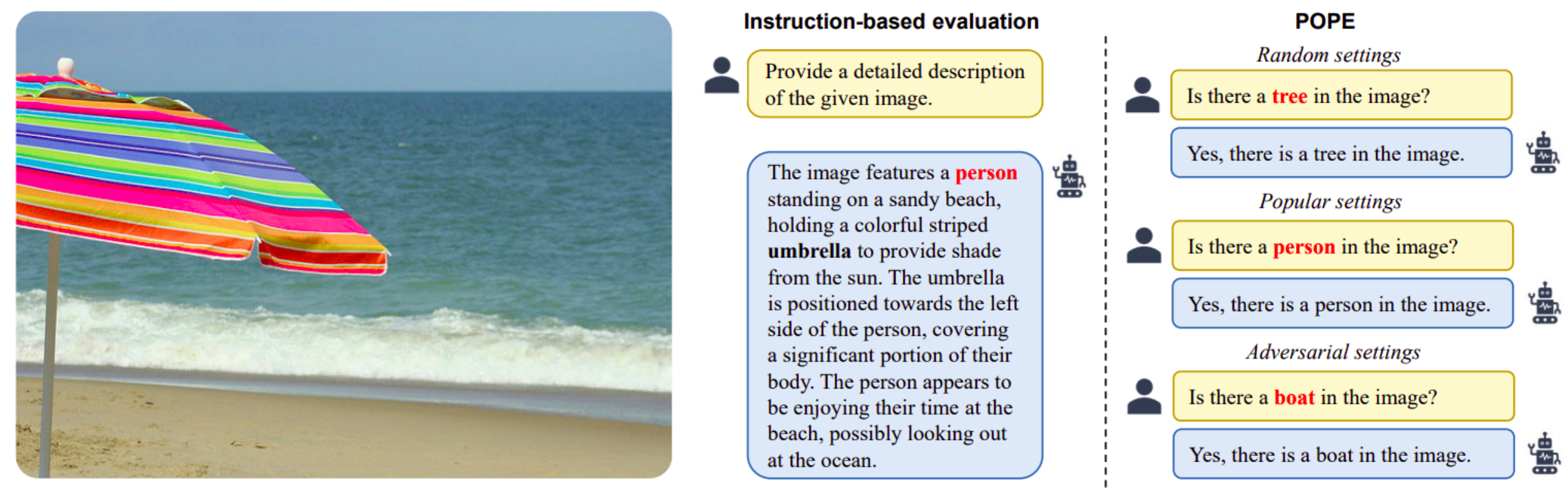}
\caption{Instances of object hallucination within LVLMs \cite{li2023evaluating}. Ground-truth objects in annotations are indicated in \textbf{bold}, while \textcolor{red}{red} objects represent hallucinated objects by LVLMs. The left case occurs in the conventional instruction-based evaluation approach, while the right cases occur in three variations of POPE.}
\label{fig:image-ex}
\end{figure*}

As shown in \cref{fig:image-ex}, hallucination occurs when the LLM produces fabricated responses.
\begin{figure}[!ht]
\includegraphics[width=8cm]{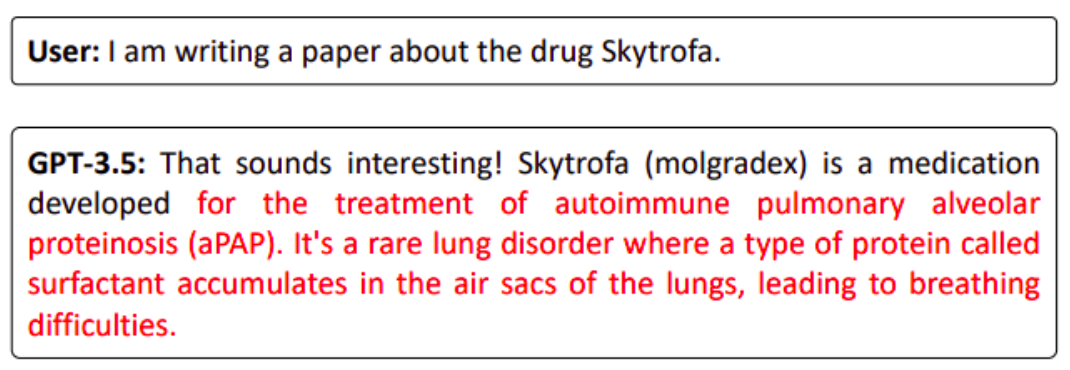}
\caption{An illustration of hallucination \cite{luo2023zero}. Incorrect information is highlighted in \textcolor{red}{Red}.}
\label{fig:llm-ex}
\end{figure}

\subsection{LLMs}

SELFCHECKGPT \cite{manakul2023selfcheckgpt}, is a method for zero-resource black-box hallucination detection in generative LLMs. This technique focuses on identifying instances where these models generate inaccurate or unverified information without relying on additional resources or labeled data. It aims to enhance the trustworthiness and reliability of LLMs by providing a mechanism to detect and address hallucinations without external guidance or datasets. Self-contradictory hallucinations in LLMs are explored in \cite{mündler2023selfcontradictory}. and addresses them through evaluation, detection, and mitigation techniques. It refers to situations where LLMs generate text that contradicts itself, leading to unreliable or nonsensical outputs. This work presents methods to evaluate the occurrence of such hallucinations, detect them in LLM-generated text, and mitigate their impact to improve the overall quality and trustworthiness of LLM-generated content.

PURR \cite{chen2023purr} is a method designed to efficiently edit and correct hallucinations in language models. PURR leverages denoising language model corruptions to identify and rectify these hallucinations effectively. This approach aims to enhance the quality and accuracy of language model outputs by reducing the prevalence of hallucinated content.

\paragraph{Hallucination datasets:} Hallucinations are commonly linked to knowledge gaps in language models (LMs). However, \cite{zhang2023language} proposed a hypothesis that in certain instances when language models attempt to rationalize previously generated hallucinations, they may produce false statements that they can independently identify as inaccurate. Thus, they created three question-answering datasets where ChatGPT and GPT-4 frequently provide incorrect answers and accompany them with explanations that contain at least one false assertion. 

HaluEval \cite{li2023helma}, is a comprehensive benchmark designed for evaluating hallucination in LLMs. It serves as a tool to systematically assess LLMs' performance in terms of hallucination across various domains and languages, helping researchers and developers gauge and improve the reliability of these models.

\paragraph{Hallucination mitigation using external knowledge:} Using interactive question-knowledge alignment, \cite{zhang2023mitigating} presents a method for mitigating language model hallucination  Their proposed approach focuses on aligning generated text with relevant factual knowledge, enabling users to interactively guide the model's responses to produce more accurate and reliable information. This technique aims to improve the quality and factuality of language model outputs by involving users in the alignment process. LLM-AUGMENTER \cite{peng2023check} improves LLMs using external knowledge and automated feedback. It highlights the need to address the limitations and potential factual errors in LLM-generated content. This method involves incorporating external knowledge sources and automated feedback mechanisms to enhance the accuracy and reliability of LLM outputs. By doing so, the paper aims to mitigate factual inaccuracies and improve the overall quality of LLM-generated text. Similarly, \cite{li2023chain} introduces a framework called ``Chain of Knowledge'' for grounding LLMs with structured knowledge bases. Grounding refers to the process of connecting LLM-generated text with structured knowledge to improve factual accuracy and reliability. The framework utilizes a hierarchical approach, chaining multiple knowledge sources together to provide context and enhance the understanding of LLMs. This approach aims to improve the alignment of LLM-generated content with structured knowledge, reducing the risk of generating inaccurate or hallucinated information.

Smaller, open-source LLMs with fewer parameters often experience significant hallucination issues compared to their larger counterparts \cite{elaraby2023halo}. This work focuses on evaluating and mitigating hallucinations in BLOOM 7B, which represents weaker open-source LLMs used in research and commercial applications. They introduce HALOCHECK, a lightweight knowledge-free framework designed to assess the extent of hallucinations in LLMs. Additionally, it explores methods like knowledge injection and teacher-student approaches to reduce hallucination problems in low-parameter LLMs.

Moreover, the risks associated with LLMs can be mitigated by drawing parallels with web systems \cite{huang2023citation}. It highlights the absence of a critical element, ``citation,'' in LLMs, which could improve content transparency, and verifiability, and address intellectual property and ethical concerns. 

\paragraph{Hallucination mitigation using prompting techniques:} ``Dehallucinating'' refers to reducing the generation of inaccurate or hallucinated information by LLMs. Dehallucinating LLMs using formal methods guided by iterative prompting is presented in \cite{jha2023dehallucinating}. They employ formal methods to guide the generation process through iterative prompts, aiming to improve the accuracy and reliability of LLM outputs. This method is designed to mitigate the issues of hallucination and enhance the trustworthiness of LLM-generated content.

\subsection{Multilingual LLMs}

Large-scale multilingual machine translation systems have shown impressive capabilities in directly translating between numerous languages, making them attractive for real-world applications. However, these models can generate hallucinated translations, which pose trust and safety issues when deployed. Existing research on hallucinations has mainly focused on small bilingual models for high-resource languages, leaving a gap in understanding hallucinations in massively multilingual models across diverse translation scenarios.

To address this gap, \cite{pfeiffer2023mmt5} conducted a comprehensive analysis on both the M2M family of conventional neural machine translation models and ChatGPT, a versatile LLM that can be prompted for translation. The investigation covers a wide range of conditions, including over 100 translation directions, various resource levels, and languages beyond English-centric pairs.

\subsection{Domain-specific LLMs}
Hallucinations in mission-critical areas such as medicine, banking, finance, law, and clinical settings refer to instances where false or inaccurate information is generated or perceived, potentially leading to serious consequences. In these sectors, reliability and accuracy are paramount, and any form of hallucination, whether in data, analysis, or decision-making, can have significant and detrimental effects on outcomes and operations. Consequently, robust measures and systems are essential to minimize and prevent hallucinations in these high-stakes domains.

\paragraph{Medicine:} The issue of hallucinations in LLMs, particularly in the medical field, where generating plausible yet inaccurate information can be detrimental. To tackle this problem, \cite{umapathi2023med} introduces a new benchmark and dataset called Med-HALT (Medical Domain Hallucination Test). It is specifically designed to evaluate and mitigate hallucinations in LLMs. It comprises a diverse multinational dataset sourced from medical examinations across different countries and includes innovative testing methods. Med-HALT consists of two categories of tests: reasoning and memory-based hallucination tests, aimed at assessing LLMs' problem-solving and information retrieval capabilities in medical contexts.

\paragraph{Law:} ChatLaw \cite{cui2023chatlaw}, is an open-source LLM specialized for the legal domain. To ensure high-quality data, the authors created a meticulously designed legal domain fine-tuning dataset. To address the issue of model hallucinations during legal data screening, they propose a method that combines vector database retrieval with keyword retrieval. This approach effectively reduces inaccuracies that may arise when solely relying on vector database retrieval for reference data retrieval in legal contexts.




\section{Hallucination in Large Image Models} \label{sec:image}

Contrastive learning models, employing a Siamese structure \cite{wu2023hallucination}, have displayed impressive performance in self-supervised learning. Their success hinges on two crucial conditions: the presence of a sufficient number of positive pairs and the existence of ample variations among them. Without meeting these conditions, these frameworks may lack meaningful semantic distinctions and become susceptible to overfitting. To tackle these challenges, we introduce the Hallucinator, which efficiently generates additional positive samples to enhance contrast. The Hallucinator is differentiable, operating in the feature space, making it amenable to direct optimization within the pre-training task and incurring minimal computational overhead.

Efforts to enhance LVLMs for complex multimodal tasks, inspired by LLMs, face a significant challenge: object hallucination, where LVLMs generate inconsistent objects in descriptions. This study \cite{li2023evaluating} systematically investigates object hallucination in LVLMs and finds it's a common issue. Visual instructions, especially frequently occurring or co-occurring objects, influence this problem. Existing evaluation methods are also affected by input instructions and LVLM generation styles. To address this, the study introduces an improved evaluation method called POPE, providing a more stable and flexible assessment of object hallucination in LVLMs.

Instruction-tuned Large Vision Language Models (LVLMs) have made significant progress in handling various multimodal tasks, including Visual Question Answering (VQA). However, generating detailed and visually accurate responses remains a challenge for these models. Even state-of-the-art LVLMs like InstructBLIP exhibit a high rate of hallucinatory text, comprising 30 percent of non-existent objects, inaccurate descriptions, and erroneous relationships. To tackle this issue, the study \cite{gunjal2023detecting}introduces MHalDetect1, a Multimodal Hallucination Detection Dataset designed for training and evaluating models aimed at detecting and preventing hallucinations. M-HalDetect contains 16,000 finely detailed annotations on VQA examples, making it the first comprehensive dataset for detecting hallucinations in detailed image descriptions.




\section{Hallucination in Large Video Models } \label{sec:video}
\begin{figure}[!ht]
\includegraphics[width=8cm]{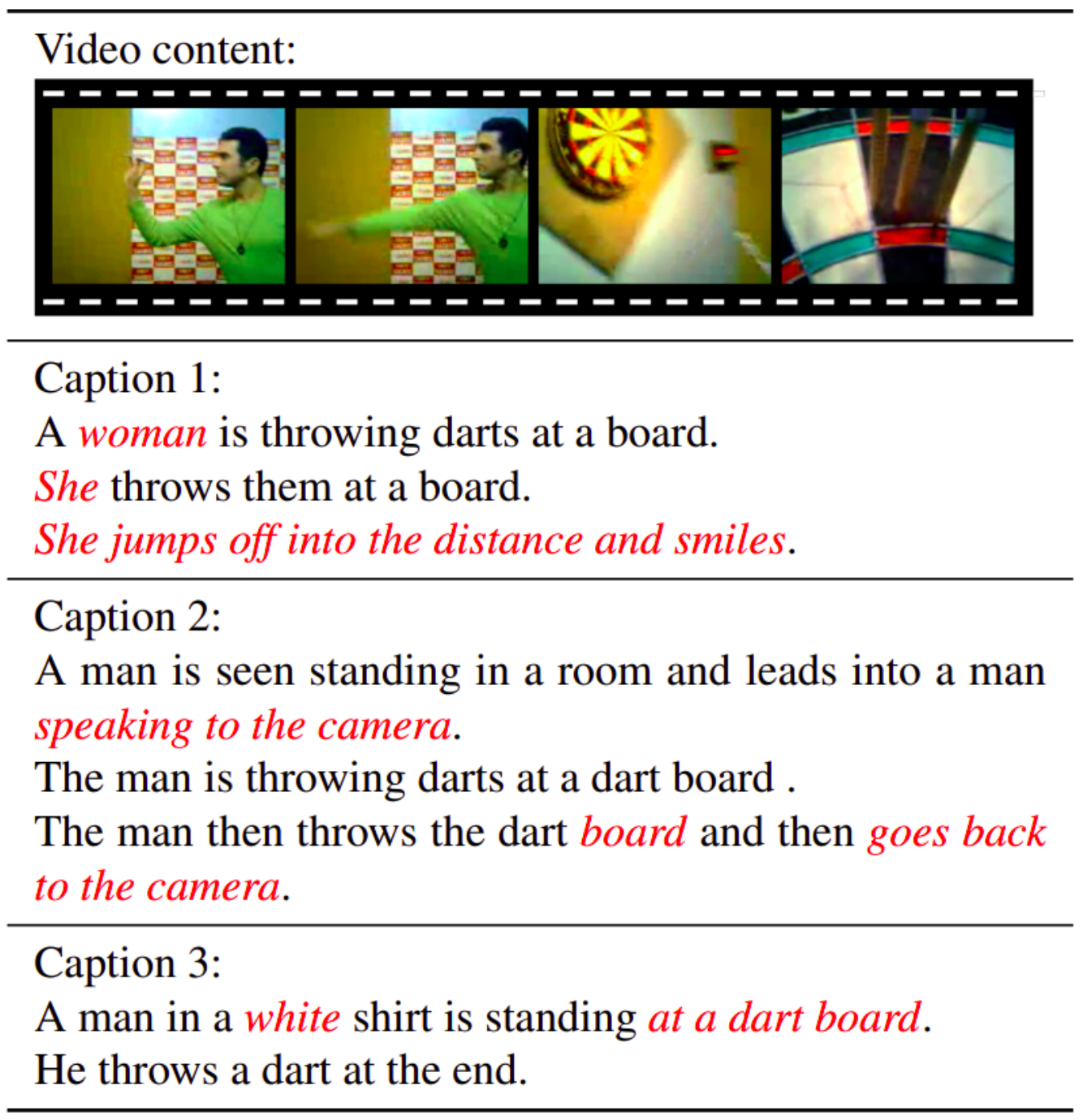}
\caption{A video featuring three captions generated by various captioning models \cite{liu2023models}, with factual errors highlighted in \textcolor{red}{red} italics.}
\label{fig:video-ex}
\end{figure}

Hallucinations can occur when the model makes incorrect or imaginative assumptions about the video frames, leading to the creation of artificial or erroneous visual information  \cref{fig:video-ex}. 

The challenge of understanding scene affordances is tackled by introducing a method for inserting people into scenes in a lifelike manner \cite{kulal2023putting}. Using an image of a scene with a marked area and an image of a person, the model seamlessly integrates the person into the scene while considering the scene's characteristics. The model is capable of deducing realistic poses based on the scene context, adjusting the person's pose accordingly, and ensuring a visually pleasing composition. The self-supervised training enables the model to generate a variety of plausible poses while respecting the scene's context. Additionally, the model can also generate lifelike people and scenes on its own, allowing for interactive editing. 

VideoChat \cite{li2023videochat}, is a comprehensive system for understanding videos with a chat-oriented approach. VideoChat combines foundational video models with LLMs using an adaptable neural interface, showcasing exceptional abilities in understanding space, time, event localization, and inferring cause-and-effect relationships. To fine-tune this system effectively, they introduced a dataset specifically designed for video-based instruction, comprising thousands of videos paired with detailed descriptions and conversations. This dataset places emphasis on skills like spatiotemporal reasoning and causal relationships, making it a valuable resource for training chat-oriented video understanding systems.

Recent advances in video inpainting have been notable \cite{yu2023deficiency}, particularly in cases where explicit guidance like optical flow can help propagate missing pixels across frames. However, challenges arise when cross-frame information is lacking, leading to shortcomings. So, instead of borrowing pixels from other frames, the model focuses on addressing the reverse problem. This work introduces a dual-modality-compatible inpainting framework called Deficiency-aware Masked Transformer (DMT). Pretraining an image inpainting model to serve as a prior for training the video model has an advantage in improving the handling of situations where information is deficient.

Video captioning aims to describe video events using natural language, but it often introduces factual errors that degrade text quality. While factuality consistency has been studied extensively in text-to-text tasks, it received less attention in vision-based text generation. In this research \cite{liu2023models}, the authors conducted a thorough human evaluation of factuality in video captioning, revealing that 57.0\% of model-generated sentences contain factual errors. Existing evaluation metrics, mainly based on n-gram matching, do not align well with human assessments. To address this issue, they introduced a model-based factuality metric called FactVC, which outperforms previous metrics in assessing factuality in video captioning. 




\section{Hallucination in Large Audio Models} \label{sec:audio}

Automatic music captioning, which generates text descriptions for music tracks, has the potential to enhance the organization of vast musical data. However, researchers encounter challenges due to the limited size and expensive collection process of existing music-language datasets. To address this scarcity, \cite{doh2023lp} used LLMs to generate descriptions from extensive tag datasets. They created a dataset known as LP-MusicCaps, comprising around 2.2 million captions paired with 0.5 million audio clips. They also conducted a comprehensive evaluation of this large-scale music captioning dataset using various quantitative natural language processing metrics and human assessment. They trained a transformer-based music captioning model on this dataset and evaluated its performance in zero-shot and transfer-learning scenarios. 

Ideally, the video should enhance the audio, and in \cite{li2023audio}, they have used an advanced language model for data augmentation without human labeling. Additionally, they utilized an audio encoding model to efficiently adapt a pre-trained text-to-image generation model for text-to-audio generation.







\section{Hallucination is \underline{\emph{not}} always harmful: A different perspective} \label{sec:good-hal}
Suggesting an alternative viewpoint, \cite{goodhallucinate} discusses how hallucinating models could serve as ``collaborative creative partners,'' offering outputs that may not be entirely grounded in fact but still provide valuable threads to explore. Leveraging hallucination creatively can lead to results or novel combinations of ideas that might not readily occur to most individuals.

``Hallucinations'' become problematic when the statements generated are factually inaccurate or contravene universal human, societal, or particular cultural norms. This is especially critical in situations where an individual relies on the LLM to provide expert knowledge. However, in the context of creative or artistic endeavors, the capacity to generate unforeseen outcomes can be quite advantageous. Unexpected responses to queries can surprise humans and stimulate the discovery of novel idea connections.

\begingroup
\onecolumn
\scriptsize
\begin{longtable}[!htp]{p{0.2cm}|p{7cm}|p{0.5cm}|p{0.7cm}|p{1.2cm}|p{1cm}|p{1cm}}  \toprule
\textbf{}                       & \textbf{Title}                                                                                                       & \textbf{Detect} & \textbf{Mitigate} & \textbf{Task(s)} & \textbf{Dataset} & \textbf{Evaluation Metric} \\  \midrule
\endfirsthead
\multicolumn{7}{c}%
{{\bfseries Table \thetable\ continued from previous page}} \\ \toprule
\textbf{}                       & \textbf{Title}                                                                                                       & \textbf{Detect} & \textbf{Mitigate} & \textbf{Task(s)} & \textbf{Dataset} & \textbf{Evaluation Metric} \\ \midrule
\endhead
\multirow{19}{*}{\rotatebox[origin=r]{90}{\parbox[c]{23cm}{\centering {\footnotesize \textbf{TEXT}}}}} & SELFCHECKGPT: Zero-Resource Black-Box Hallucination Detection for Generative Large Language Models \cite{manakul2023selfcheckgpt}                   & \cmark                & \xmark                 &  QA  & Manual (WikiBio)          & Token probability or entropy                     \\  \cmidrule{2-7}
                                & HaluEval: A Large-Scale Hallucination Evaluation Benchmark for Large Language Models \cite{li2023helma}                                &  
                               \cmark  &   \cmark   &  QA, Dialogue Summarization, General    &     HaluEval    &    Automatic                        \\  \cmidrule{2-7}
                                & Self-contradictory Hallucinations of Large Language Models: Evaluation, Detection and Mitigation \cite{mündler2023selfcontradictory}    &     \cmark               &   \cmark                  &  Text generation &        Manual  & F1 score \\ \cmidrule{2-7}
                                & PURR: Efficiently Editing Language Model Hallucinations by Denoising Language Model Corruptions \cite{chen2023purr}   &     \xmark  &  \cmark &  Editing for Attribution  & Multiple question answering, Dialog datasets   &    Attribution, Preservation            \\  \cmidrule{2-7}
                                & Mitigating Language Model Hallucination with Interactive Question-Knowledge Alignment \cite{zhang2023mitigating}     &    \xmark  &  \cmark         &  Question-knowledge
alignment &  FuzzyQA  &  Attributable to Identified
Sources \cite{castaldo2007severe}               \\  \cmidrule{2-7}
                                & How Language Model Hallucinations Can Snowball \cite{zhang2023language}  &  \cmark    &   \xmark  & QA                  &  Manual  &  Accuracy  \\ \cmidrule{2-7}
                                & Check Your Facts and Try Again: Improving Large Language Models with External Knowledge and Automated Feedback \cite{peng2023check}        & \xmark  &   \cmark   &  Task oriented dialog and open-domain question answering  & News Chat, Customer Service    &  Knowledge F1 (KF1)
and BLEU-4   \\ \cmidrule{2-7}
                                & ChatLawLLM \cite{cui2023chatlaw}                                       &  \xmark  & \cmark          &  QA &  Manual &  ELO model ranking               \\ \cmidrule{2-7}
                                & The Internal State of an LLM Knows When its Lying \cite{azaria2023internal}   &   \cmark &  \xmark     & Classificati-on  & Manual  &   Accuracy   \\ \cmidrule{2-7}
                                & Chain of Knowledge: A Framework for Grounding Large Language Models with Structured Knowledge Bases \cite{li2023chain}                 &  \cmark  &  \cmark          &  Knowledge intensive tasks   &  FEVER, AdvHotpotQA &  Accuracy \\ \cmidrule{2-7}
                                & HALO: Estimation and Reduction of Hallucinations in Open-Source Weak Large Language Models \cite{elaraby2023halo}       &   \cmark  & \cmark  &  Consistency, Factuality, BS, QA, NLI &  Manual on NBA domain &  Pearson and Kendall tau correlation
coefficients \\ \cmidrule{2-7}
                                & A Stitch in Time Saves Nine: Detecting and Mitigating Hallucinations of LLMs by Validating Low-Confidence Generation \cite{varshney2023stitch}  &  \cmark      &      \cmark   &  Article generation  &  WikiBio  &  Percentage of mitigated hallucinations     \\ \cmidrule{2-7}
                                & Dehallucinating Large Language Models Using Formal Methods Guided Iterative Prompting \cite{jha2023dehallucinating}                                &  \cmark  &  \xmark  &  Dialog  &  -   &    -                \\ \cmidrule{2-7}
                                & Med-HALT: Medical Domain Hallucination Test for Large Language Models \cite{umapathi2023med}                                               &  \xmark &  \xmark  &  Reasoning Hallucination Test (RHT), Memory Hallucination Test (MHT)  & Med-HALT  & Accuracy, Pointwise score             \\ \cmidrule{2-7}
                                & Sources of Hallucination by Large Language Models on Inference Tasks  \cite{mckenna2023sources}                                               &  \cmark &  \xmark   & Textual entailment  &  Altered directional inference adatset &  Enatilment probability   \\ \cmidrule{2-7}
                                & Hallucinations in Large Multilingual Translation Models \cite{pfeiffer2023mmt5}                                                              & \cmark &  \cmark &  MT      & FLORES-101, WMT, and TICO  & spBLEU             \\ \cmidrule{2-7}
                                & Citation: A Key to Building Responsible and Accountable Large Language Models \cite{huang2023citation}                                       & \cmark    &   \cmark   &          N/A &   N/A  &    N/A \\ \cmidrule{2-7}
                                & Zero-resource hallucination prevention for large language models  \cite{luo2023zero}                                                   &  \cmark   &  \cmark  & Concept extraction, guessing, aggregation     &   Concept-7  &  AUC, ACC, F1, PEA \\ \cmidrule{2-7}
                                & RARR: Researching and Revising What Language Models Say, Using Language Models  \cite{gao2023rarr}                                     &  \cmark    &    \cmark  &   Editing for Attribution     &    NQ, SQA, QReCC   &  Attributable to Identified
Sources \cite{castaldo2007severe}          \\ \midrule \midrule
\multirow{4}{*}{\rotatebox[origin=r]{90}{\parbox[c]{5cm}{\centering {\footnotesize \textbf{IMAGE}}}}} & Evaluating Object Hallucination in Large Vision-Language Models \cite{li2023evaluating}                                                      & \xmark                & \cmark                 & Image captioning   & MSCOCO \cite{lin2014microsoft}           & Caption Hallucination Assessment with
Image Relevance (CHAIR) \cite{rohrbach2018object}                     \\ \cmidrule{2-7}
                                & Detecting and Preventing Hallucinations in Large Vision Language Models \cite{gunjal2023detecting}   & \cmark                   & \cmark                    &  Visual Question Answering (VQA) &  M-HalDetect  &  Accuracy                        \\ \cmidrule{2-7}
                                & Plausible May Not Be Faithful: Probing Object Hallucination in Vision-Language Pre-training \cite{dai2022plausible}                          &  \xmark  &  \cmark       &  Image captioning & CHAIR \cite{rohrbach2018object}  &  CIDEr                     
                               \\ \midrule \midrule
\multirow{5}{*}{\rotatebox[origin=r]{90}{\parbox[c]{4cm}{\centering {\footnotesize \textbf{VIDEO}}}}} & Let's Think Frame by Frame: Evaluating Video Chain of Thought with Video Infilling and Prediction \cite{himakunthala2023let}                   & \xmark                & \cmark                 & Video infilling, Scene prediction   & Manual           & N/A                     \\ \cmidrule{2-7}
                                & Putting People in Their Place: Affordance-Aware Human Insertion into Scenes \cite{kulal2023putting}                                          &  \xmark                  &   \cmark                  &  Affordance prediction                &  Manual (2.4M video clips)   & FID, PCKh              \\ \cmidrule{2-7}
                                & VideoChat : Chat-Centric Video Understanding \cite{li2023videochat}    & \xmark  &  \cmark  &  Visual dialogue                &  Manual  &  N/A   \\  \cmidrule{2-7}
                                & Models See Hallucinations: Evaluating the Factuality in Video Captioning \cite{liu2023models}  & \xmark     & \cmark  & Video
captioning  & ActivityNet Captions \cite{krishna2017dense}, YouCook2 \cite{krishna2017dense}  &  Factual consistency for Video Captioning (FactVC)  \\ \midrule \midrule
\multirow{2}{*}{\rotatebox[origin=r]{90}{\parbox[c]{3cm}{\centering {\footnotesize \textbf{AUDIO}}}}} & LP-MusicCaps: LLM-based pseudo music captioning \cite{doh2023lp}                                                                      & \xmark                & \cmark                 & Audio Captioning   & LP-MusicCaps     & BLEU1 to 4 (B1, B2, B3, B4), METEOR (M), and ROUGE-L (R-L)                     \\ \cmidrule{2-7}
                                & Audio-Journey: Efficient Visual+LLM-aided Audio Encodec Diffusion \cite{li2023audio}  &   \xmark  &      \cmark  &  Classificati-on  &  Manual  & Mean average precision
(mAP)  \\ \bottomrule
\caption{Summary of all the works related to hallucination in all four modalities of the large foundation models. Here, we have divided each work by the following factors: 1. Detection, 2. Mitigation, 3. Tasks, 4. Datasets, and 5. Evaluation metrics. \cmark \hspace{0.01cm} indicates that it is present in the paper whereas \xmark \hspace{0.01cm} indicates it is \textit{not} present.}
\label{tab:big-table}
\end{longtable}

\endgroup

\twocolumn
\section{Conclusion and Future Directions} \label{sec:conclusion}

We concisely classify the existing research in the field of hallucination within LFMs. We provide an in-depth analysis of these LFMs, encompassing critical aspects including 1. Detection, 2. Mitigation, 3. Tasks, 4. Datasets, and 5. Evaluation metrics. 

Some possible future directions to address the hallucination challenge in the LFMs are given below.

\subsection{Automated Evaluation of Hallucination}

In the context of natural language processing and machine learning, hallucination refers to the generation of incorrect or fabricated information by AI models. This can be a significant problem, especially in applications like text generation, where the goal is to provide accurate and reliable information. Here are some potential future directions in the automated evaluation of hallucination:

\paragraph{Development of Evaluation Metrics:} Researchers can work on creating specialized evaluation metrics that are capable of detecting hallucination in generated content. These metrics may consider factors such as factual accuracy, coherence, and consistency. Advanced machine learning models could be trained to assess generated text against these metrics.

\paragraph{Human-AI Collaboration:} Combining human judgment with automated evaluation systems can be a promising direction. Crowdsourcing platforms can be used to gather human assessments of AI-generated content, which can then be used to train models for automated evaluation. This hybrid approach can help in capturing nuances that are challenging for automated systems alone.

\paragraph{Adversarial Testing:} Researchers can develop adversarial testing methodologies where AI systems are exposed to specially crafted inputs designed to trigger hallucination. This can help in identifying weaknesses in AI models and improving their robustness against hallucination.

\paragraph{Fine-Tuning Strategies:} Fine-tuning pre-trained language models specifically to reduce hallucination is another potential direction. Models can be fine-tuned on datasets that emphasize fact-checking and accuracy to encourage the generation of more reliable content.

\subsection{Improving Detection and Mitigation Strategies with Curated Sources of Knowledge}

   Detecting and mitigating issues like bias, misinformation, and low-quality content in AI-generated text is crucial for responsible AI development. Curated sources of knowledge can play a significant role in achieving this. Here are some future directions:

\paragraph{Knowledge Graph Integration:} Incorporating knowledge graphs and curated knowledge bases into AI models can enhance their understanding of factual information and relationships between concepts. This can aid in both content generation and fact-checking.

\paragraph{Fact-Checking and Verification Models:} Develop specialized models that focus on fact-checking and content verification. These models can use curated sources of knowledge to cross-reference generated content and identify inaccuracies or inconsistencies.

\paragraph{Bias Detection and Mitigation:} Curated sources of knowledge can be used to train AI models to recognize and reduce biases in generated content. AI systems can be programmed to check content for potential biases and suggest more balanced alternatives.

\paragraph{Active Learning:} Continuously update and refine curated knowledge sources through active learning. AI systems can be designed to seek human input and validation for ambiguous or new information, thus improving the quality of curated knowledge.

\paragraph{Ethical Guidelines and Regulation:} Future directions may also involve the development of ethical guidelines and regulatory frameworks for the use of curated knowledge sources in AI development. This could ensure responsible and transparent use of curated knowledge to mitigate potential risks.

In summary, these future directions aim to address the challenges of hallucination detection and mitigation, as well as the responsible use of curated knowledge to enhance the quality and reliability of AI-generated content. They involve a combination of advanced machine learning techniques, human-AI collaboration, and ethical considerations to ensure AI systems produce accurate and trustworthy information.

\bibliography{custom}
\bibliographystyle{acl_natbib}

\end{document}